\documentclass[runningheads]{llncs}
\usepackage{graphicx}
\usepackage{amsmath,amssymb} 
\usepackage{color}
\usepackage[width=122mm,left=12mm,paperwidth=146mm,height=193mm,top=12mm,paperheight=217mm]{geometry}
\usepackage{epstopdf}

\newcommand{\SEGNET}{temporal segment network}

\begin{document}
\pagestyle{headings}
\mainmatter

\title{Temporal Segment Networks: Towards Good Practices for Deep Action Recognition} 

\titlerunning{TSNs: Towards Good Practices for Deep Action Recognition}

\authorrunning{L. Wang \and Y. Xiong \and Z. Wang \and Y. Qiao \and D. Lin  \and X. Tang \and L. Van Gool}

\author{
	Limin Wang\textsuperscript{1}\and 
	Yuanjun Xiong\textsuperscript{2}\and 
	Zhe Wang\textsuperscript{3}\and 
	Yu Qiao\textsuperscript{3}\and 
	Dahua Lin\textsuperscript{2}\and \\
	Xiaoou Tang\textsuperscript{2}\and
	Luc Van Gool\textsuperscript{1} 
}


\institute{
    \textsuperscript{1}Computer Vision Lab, ETH Zurich, Switzerland\\
	\textsuperscript{2}Department of Information Engineering, The Chinese University of Hong Kong\\
	\textsuperscript{3}Shenzhen Institutes of Advanced Technology, CAS, China\\
}

\maketitle

\begin{abstract}
	Deep convolutional networks have achieved great success for visual recognition in still images. However, for action recognition in videos, the advantage over traditional methods is not so evident. This paper aims to discover the principles to design effective ConvNet architectures for action recognition in videos and learn these models given limited training samples. 
	Our first contribution is \SEGNET~(TSN), a novel framework for video-based action recognition. which is based on the idea of long-range temporal structure modeling.
	It combines a sparse temporal sampling strategy and video-level supervision to enable efficient and effective learning using the whole action video.
	The other contribution is our study on a series of good practices in learning ConvNets on video data with the help of \SEGNET.
	Our approach obtains the state-the-of-art performance on the datasets of HMDB51 ($ 69.4\% $) and UCF101 ($ 94.2\% $). We also visualize the learned ConvNet models, which qualitatively demonstrates the effectiveness of \SEGNET~and the proposed good practices. \footnote{Models and code at \url{https://github.com/yjxiong/temporal-segment-networks}.}
\end{abstract}

\section{Introduction}\label{sec:intro}

Video-based action recognition has drawn a significant amount of attention from the academic community~\cite{SimonyanZ14,WangS13a,WangQT13a,Ng15,WangQT15a,gan2015devnet}, owing to its applications in many areas like security and behavior analysis.
In action recognition, there are two crucial and complementary aspects: appearances and dynamics.
The performance of a recognition system depends, to a large extent, on whether it is able to extract and utilize relevant information therefrom.
However, extracting such information is non-trivial due to a number of complexities, such as scale variations, view point changes, and camera motions.
Thus it becomes crucial to design effective representations that can deal with these challenges while preserve categorical information of action classes.
Recently, Convolutional Networks (ConvNets)~\cite{lecun-98} have witnessed great success in classifying images of objects, scenes, and complex events~\cite{KrizhevskySH12,SimonyanZ14a,SzegedyLJSRAEVR14,XiongY2015}.
ConvNets have also been introduced to solve the problem of video-based action recognition~\cite{KarpathyTSLSF14,SimonyanZ14,TranBFTP15,ZhangWWQW16}.
Deep ConvNets come with great modeling capacity and are capable of learning discriminative representation from raw visual data with the help of large-scale supervised datasets. However, unlike image classification, end-to-end deep ConvNets remain unable to achieve significant advantage over traditional hand-crafted features for video-based action recognition. 

In our view, the application of ConvNets in video-based action recognition is impeded by two major obstacles.
{\em First}, long-range temporal structure plays an important role in understanding the dynamics in action videos \cite{NieblesCF10,GaidonHS13,WangQT14a,FernandoGMGT15}.
However, mainstream ConvNet frameworks~\cite{SimonyanZ14,TranBFTP15} usually focus on appearances and short-term motions, thus lacking the capacity to incorporate long-range temporal structure.
Recently there are a few attempts~\cite{varol,Ng15,DonahueJ2015} to deal with this problem.
These methods mostly rely on dense temporal sampling with a pre-defined sampling interval. This approach would incur excessive computational cost when applied to long video sequences, which limits its application in real-world practice and poses a risk of missing important information for videos longer than the maximal sequence length.
{\em Second}, in practice, training deep ConvNets requires a large volume of training samples to achieve optimal performance. 
However, due to the difficulty in data collection and annotation, publicly available action recognition datasets (e.g. UCF101~\cite{Soomro12}, HMDB51~\cite{KuehneJGPS11}) remain limited, in both size and diversity.
Consequently, very deep ConvNets~\cite{SimonyanZ14a,IoffeS15}, which have attained remarkable success in image classification, are confronted with high risk of over-fitting.

These challenges motivate us to study two problems: \emph{1) how to design an effective and efficient video-level framework for learning video representation that is able to capture long-range temporal structure; 2) how to learn the ConvNet models given limited training samples}. 
In particular, we build our method on top of the successful two-stream architecture~\cite{SimonyanZ14} while tackling the problems mentioned above.
In terms of temporal structure modeling, a key observation is that consecutive frames are highly redundant.
Therefore, dense temporal sampling, which usually results in highly similar sampled frames, is unnecessary.
Instead a sparse temporal sampling strategy will be more favorable in this case.
Motivated by this observation, we develop a video-level framework, called \emph{\SEGNET}~(TSN).
This framework extracts short snippets over a long video sequence with a sparse sampling scheme, where the samples distribute uniformly along the temporal dimension.
Thereon, a segmental structure is employed to aggregate information from the sampled snippets.
In this sense, \SEGNET s are capable of modeling long-range temporal structure over the whole video.
Moreover, this sparse sampling strategy preserves relevant information with dramatically lower cost, thus enabling end-to-end learning over long video sequences under a reasonable budget in both time and computing resources.

To unleash the full potential of \SEGNET~framework, we adopt very deep ConvNet architectures~\cite{IoffeS15,SimonyanZ14a} introduced recently, and explored a number of good practices to overcome the aforementioned difficulties caused by the limited number of training samples, including 1) cross-modality pre-training; 2) regularization; 3) enhanced data augmentation. Meanwhile, to fully utilize visual content from videos, we empirically study four types of input modalities to two-stream ConvNets, namely a single RGB image, stacked RGB difference, stacked optical flow field, and stacked warped optical flow field.

We perform experiments on two challenging action recognition datasets, namely UCF101~\cite{Soomro12}  and HMDB51~\cite{KuehneJGPS11}, to verify the effectiveness of our method. In experiments, models learned using the temporal segment network significantly outperform the state of the art on these two challenging action recognition datasets. We also visualize the our learned two-stream models trying to provide some insights for future action recognition research.

\section{Related Works}
\label{sec:rw}
Action recognition has been extensively studied in past few years~\cite{WangS13a,gan2016you,PengWWQ14,gan2016recognizing,FernandoGMGT15}. 
Previous works related to ours fall into two categories: (1) convolutional networks for action recognition, (2) temporal structure modeling.

\textbf{Convolutional Networks for Action Recognition.} 
Several works have been trying to design effective ConvNet architectures for action recognition in videos~\cite{KarpathyTSLSF14,SimonyanZ14,TranBFTP15,JiXYY13,SunJYS15}. Karpathy \emph{et al.}~\cite{KarpathyTSLSF14} tested ConvNets with deep structures on a large dataset (Sports-1M). Simonyan \emph{et al.}~\cite{SimonyanZ14} designed two-stream ConvNets containing spatial and temporal net by exploiting ImageNet dataset for pre-training and calculating optical flow to explicitly capture motion information. Tran \emph{et al.}~\cite{TranBFTP15} explored 3D ConvNets~\cite{JiXYY13} on the realistic and large-scale video datasets, where they tried to learn both appearance and motion features with 3D convolution operations. Sun \emph{et al.}~\cite{SunJYS15} proposed a factorized spatio-temporal ConvNets and exploited different ways to decompose 3D convolutional kernels. Recently, several works focused on modeling long-range temporal structure with ConvNets~\cite{Ng15,varol,DonahueJ2015}. However, these methods directly operated on a longer continuous video streams. Limited by computational cost these methods usually process sequences of fixed lengths ranging from 64 to 120 frames. It is non-trivial for these methods to learn from entire video due to their limited temporal coverage. Our method differs from these end-to-end deep ConvNets by its novel adoption of a sparse temporal sampling strategy, which enables efficient learning using the entire videos without the limitation of sequence length.

\textbf{Temporal Structure Modeling.} 
Many research works have been devoted to modeling the temporal structure for action recognition~\cite{NieblesCF10,GaidonHS13,WangQT14a,PirsiavashR14,WangQT14b,FernandoGMGT15}. Gaidon \emph{et al.}~\cite{GaidonHS13} annotated each atomic action for each video and proposed Actom Sequence Model (ASM) for action detection. Niebles \emph{et al.}~\cite{NieblesCF10} proposed to use latent variables to model the temporal decomposition of complex actions, and resorted to the Latent SVM~\cite{FelzenszwalbGMR10} to learn the model parameters in an iterative approach. Wang \emph{et al.}~\cite{WangQT14a} and Pirsiavash \emph{et al.}~\cite{PirsiavashR14} extended the temporal decomposition of complex action into a hierarchical manner using Latent Hierarchical Model (LHM) and Segmental Grammar Model (SGM), respectively. Wang \emph{et al.}~\cite{WangQT14b} designed a sequential skeleton model (SSM) to capture the relations among dynamic-poselets, and performed spatio-temporal action detection. Fernando~\cite{FernandoGMGT15} modeled the temporal evolution of BoVW representations for action recognition. 
These methods, however, remain unable to assemble an end-to-end learning scheme for modeling the temporal structure. The proposed temporal segment network, while also emphasizing this principle, is the first framework for end-to-end temporal structure modeling on the entire videos.

\section{Action Recognition with Temporal Segment Networks}
\label{sec:vds}

In this section, we give detailed descriptions of performing action recognition with \SEGNET s.
Specifically, we first introduce the basic concepts in the framework of \SEGNET. 
Then, we study the good practices in learning two-stream ConvNets within the \SEGNET~framework. Finally, we describe the testing details of the learned two-stream ConvNets.

\subsection{Temporal Segment Networks}\label{sec:seg}

\begin{figure}[t]
	\centering
	\includegraphics[width=.98\linewidth]{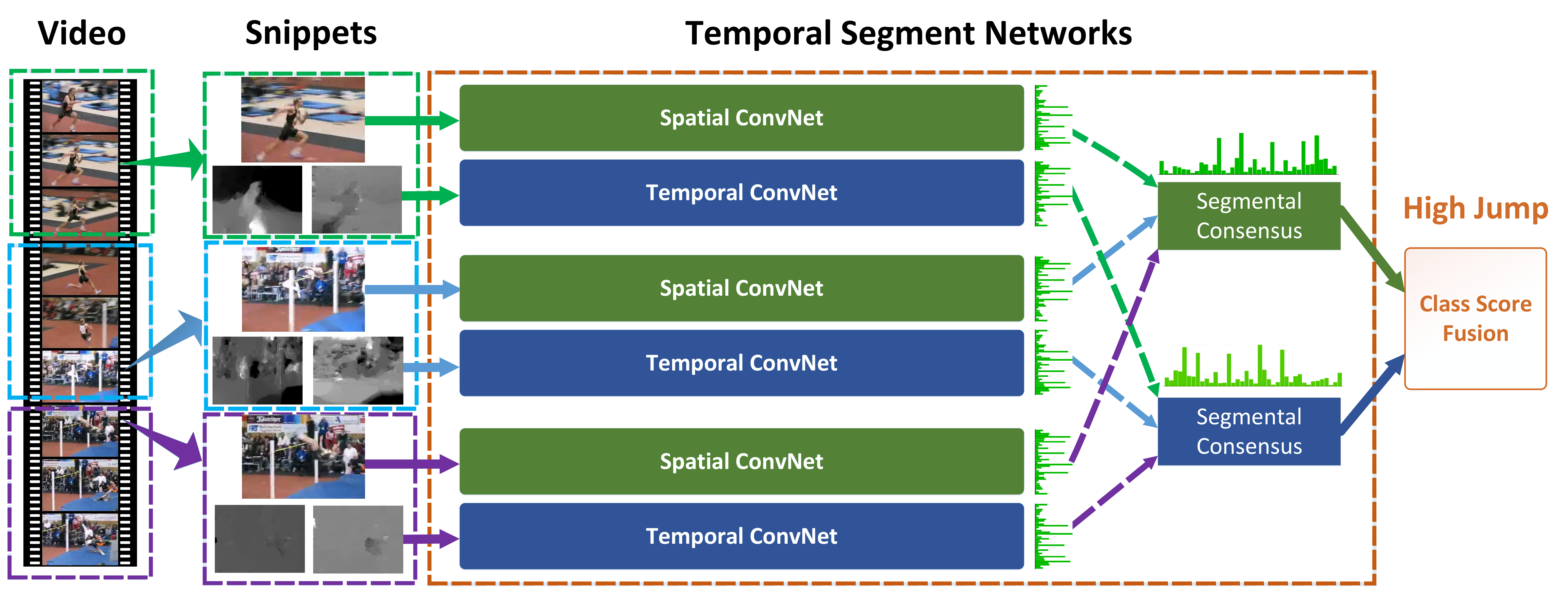}
	\caption{Temporal segment network: One input video is divided into $ K $ segments and a short snippet is randomly selected from each segment. The class scores of different snippets are fused by an the segmental consensus function to yield segmental consensus, which is a video-level prediction. Predictions from all modalities are then fused to produce the final prediction. ConvNets on all snippets share parameters.}
	\label{fig:pipeline}
\end{figure}

As we discussed in Sec.~\ref{sec:intro}, an obvious problem of the two-stream ConvNets in their current forms is their inability in modeling long-range temporal structure.
This is mainly due to their limited access to temporal context as they are designed to operate only on a single frame (spatial networks) or a single stack of frames in a short snippet (temporal network).
However, complex actions, such as sports action, comprise multiple stages spanning over a relatively long time. 
It would be quite a loss failing to utilize long-range temporal structures in these actions into ConvNet training.
To tackle this issue, we propose \SEGNET, a video-level framework as shown in Figure \ref{fig:pipeline}, to enable to model dynamics throughout the whole video.

Specifically, our proposed \SEGNET~framework, aiming to utilize the visual information of entire videos to perform video-level prediction, is also composed of spatial stream ConvNets and temporal stream ConvNets. 
Instead of working on single frames or frame stacks, \SEGNET s operate on a sequence of short snippets sparsely sampled from the entire video.
Each snippet in this sequence will produce its own preliminary prediction of the action classes.
Then a consensus among the snippets will be derived as the video-level prediction.
In the learning process, the loss values of video-level predictions, 
other than those of snippet-level predictions which were used in two-stream ConvNets, are optimized by iteratively updating the model parameters.

Formally, given a video $V$, we divide it into $K$ segments $\{S_1, S_2, \cdots, S_K\}$ of equal durations. Then, the temporal segment network models a sequence of snippets as follows:
\begin{equation}
\mathrm{TSN}(T_1, T_2, \cdots, T_K) = \mathcal{H} (\mathcal{G}(\mathcal{F}(T_1;\mathbf{W}), \mathcal{F}(T_2;\mathbf{W}), \cdots, \mathcal{F}(T_K; \mathbf{W}))).
\label{equ:model}
\end{equation}
Here $(T_1, T_2, \cdots, T_K)$ is a sequence of snippets. Each snippet $ T_k $ is randomly sampled from its corresponding segment $S_k$. $\mathcal{F}(T_k; \mathbf{W})$ is the function representing a ConvNet with parameters $\mathbf{W}$ which operates on the short snippet $T_k$ and produces class scores for all the classes.
The segmental consensus function $\mathcal{G}$ combines the outputs from multiple short snippets to obtain a consensus of class hypothesis among them. 
Based on this consensus, the prediction function $\mathcal{H}$ predicts the probability of each action class for the whole video.
Here we choose the widely used Softmax function for $\mathcal{H}$. Combining with standard categorical cross-entropy loss, the final loss function regarding the segmental consensus $ \mathbf{G} = \mathcal{G}(\mathcal{F}(T_1;\mathbf{W}), \mathcal{F}(T_2;\mathbf{W}), \cdots, \mathcal{F}(T_K; \mathbf{W})) $ is formed as
\begin{equation}
\mathcal{L}(y, \mathbf{G}) = -\sum_{i=1}^{C}y_i \left(G_i - \log\sum_{j=1}^{C}\exp{G_j}\right),
\end{equation}
where $ C $ is the number of action classes and $ y_i $ the groundtruth label concerning class $ i $.
In experiments, the number of snippets $K$ is set to $ 3 $ according to previous works on temporal modeling \cite{GaidonHS13,WangQT14a}. 
The form of consensus function $ \mathcal{G} $ remains an open question. 
In this work we use the simplest form of $ \mathcal{G} $, where $ G_i = g(\mathcal{F}_i(T_1), \ldots, \mathcal{F}_i(T_K))$. 
Here a class score $ G_i $ is inferred from the scores of the same class on all the snippets, using an aggregation function $ g $.
We empirically evaluated several different forms of the aggregation function $ g $, including evenly averaging, maximum, and weighted averaging in our experiments.
Among them, evenly averaging is used to report our final recognition accuracies.

This \SEGNET~is differentiable or at least has subgradients, depending on the choice of $ g $.
This allows us to utilize the multiple snippets to jointly optimize the model parameters $\mathbf{W}$ with standard back-propagation algorithms.
In the back-propagation process, the gradients of model parameters $ \mathbf{W} $ with respect to the loss value $\mathcal{L}$ can be derived as
\begin{equation}\label{eq:grad}
\frac{\partial \mathcal{L}(y, \mathbf{G})}{\partial \mathbf{W}} = 
\frac{\partial\mathcal{L}}{\partial \mathbf{G}} \sum_{k=1}^K \frac{\partial \mathcal{G}}{\partial \mathcal{F}(T_k)} \frac{\partial \mathcal{F}(T_k)}{\partial \mathbf{W}},
\end{equation}
where $ K $ is number of segments \SEGNET~uses.

When we use a gradient-based optimization method, like stochastic gradient descent (SGD), to learn the model parameters, Eq.~\ref{eq:grad} guarantees that the parameter updates are utilizing the segmental consensus $ \mathbf{G} $ derived from all snippet-level prediction.
Optimized in this manner, \SEGNET  can learn model parameters from the entire video rather than a short snippet. 
Meanwhile, by fixing $ K $ for all videos, we assemble a sparse temporal sampling strategy, where the sampled snippets contain only a small portion of the frames.
It drastically reduces the computational cost for evaluating ConvNets on the frames, compared with previous works using densely sampled frames~\cite{Ng15,varol,DonahueJ2015}.

\subsection{Learning Temporal Segment Networks}\label{sec:training}

Temporal segment network~provides a solid framework to perform video-level learning, but to achieve optimal performance, a few practical concerns have to be taken care of, for example the limited numberof training samples.
To this end, we study a series of good practices in training deep ConvNets on video data, which are also directly applicable in learning \SEGNET s.

\textbf{Network Architectures.}
Network architecture is an important factor in neural network design. Several works have shown that deeper structures improve object recognition performance \cite{SimonyanZ14a,SzegedyLJSRAEVR14}. However, the original two-stream ConvNets~\cite{SimonyanZ14} employed a relatively shallow network structure (ClarifaiNet \cite{ZeilerF14}).
In this work, we choose the Inception with Batch Normalization (BN-Inception)~\cite{IoffeS15} as building block, due to its good balance between accuracy and efficiency.
We adapt the original BN-Inception architecture to the design of two-stream ConvNets. Like in the original two-stream ConvNets~\cite{SimonyanZ14}, the spatial stream ConvNet operates on a single RGB images, and the temporal stream ConvNet takes a stack of consecutive optical flow fields as input. 

\textbf{Network Inputs.}
We are also interested in exploring more input modalities to enhance the discriminative power of \SEGNET s. Originally, the two-stream ConvNets used RGB images for the spatial stream and stacked optical flow fields for the temporal stream.
Here, we propose to study two extra modalities, namely \emph{RGB difference} and \emph{warped optical flow fields}.

\begin{figure}[t]
	\centering{
		\includegraphics[width=0.2\linewidth]{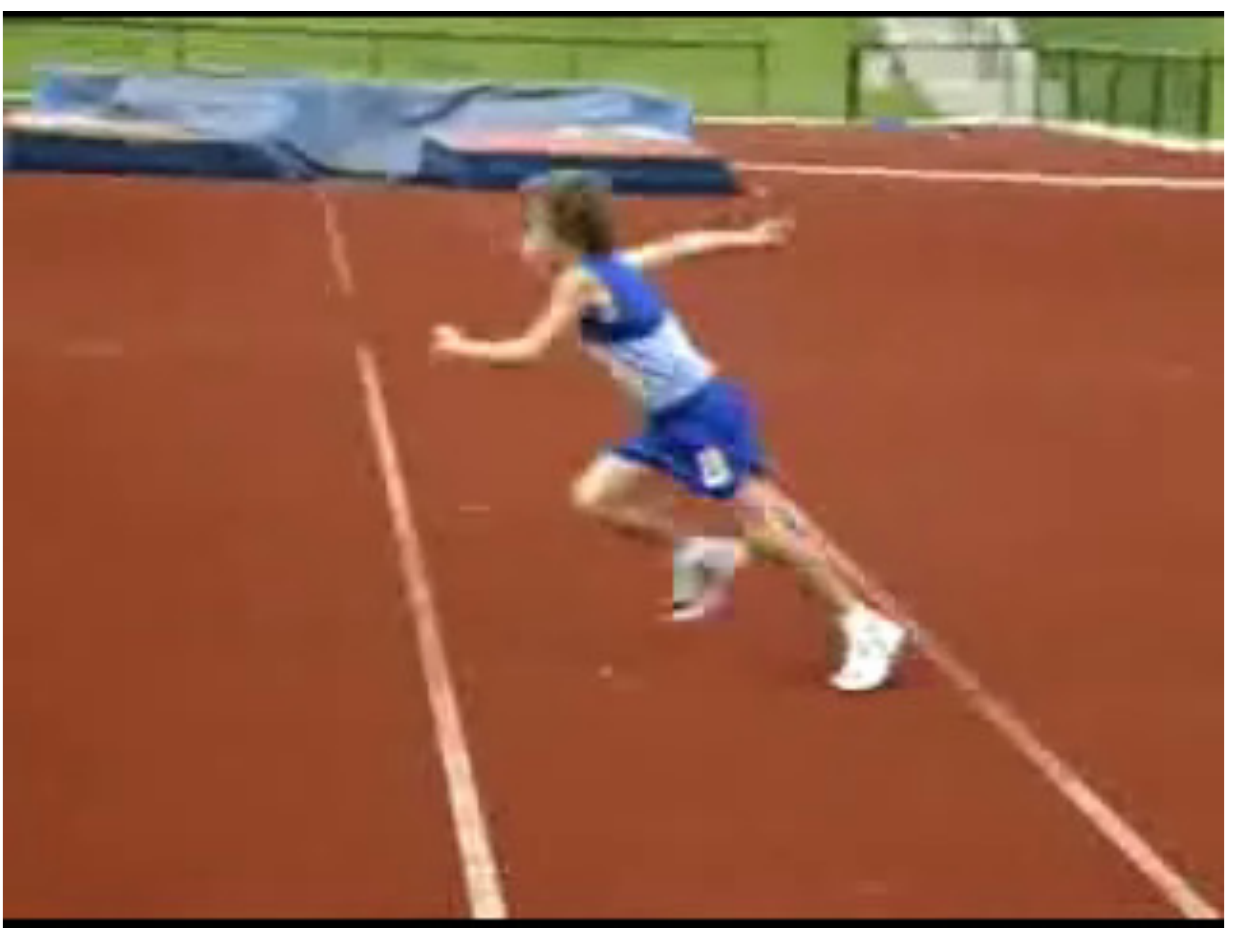}
		\includegraphics[width=0.2\linewidth]{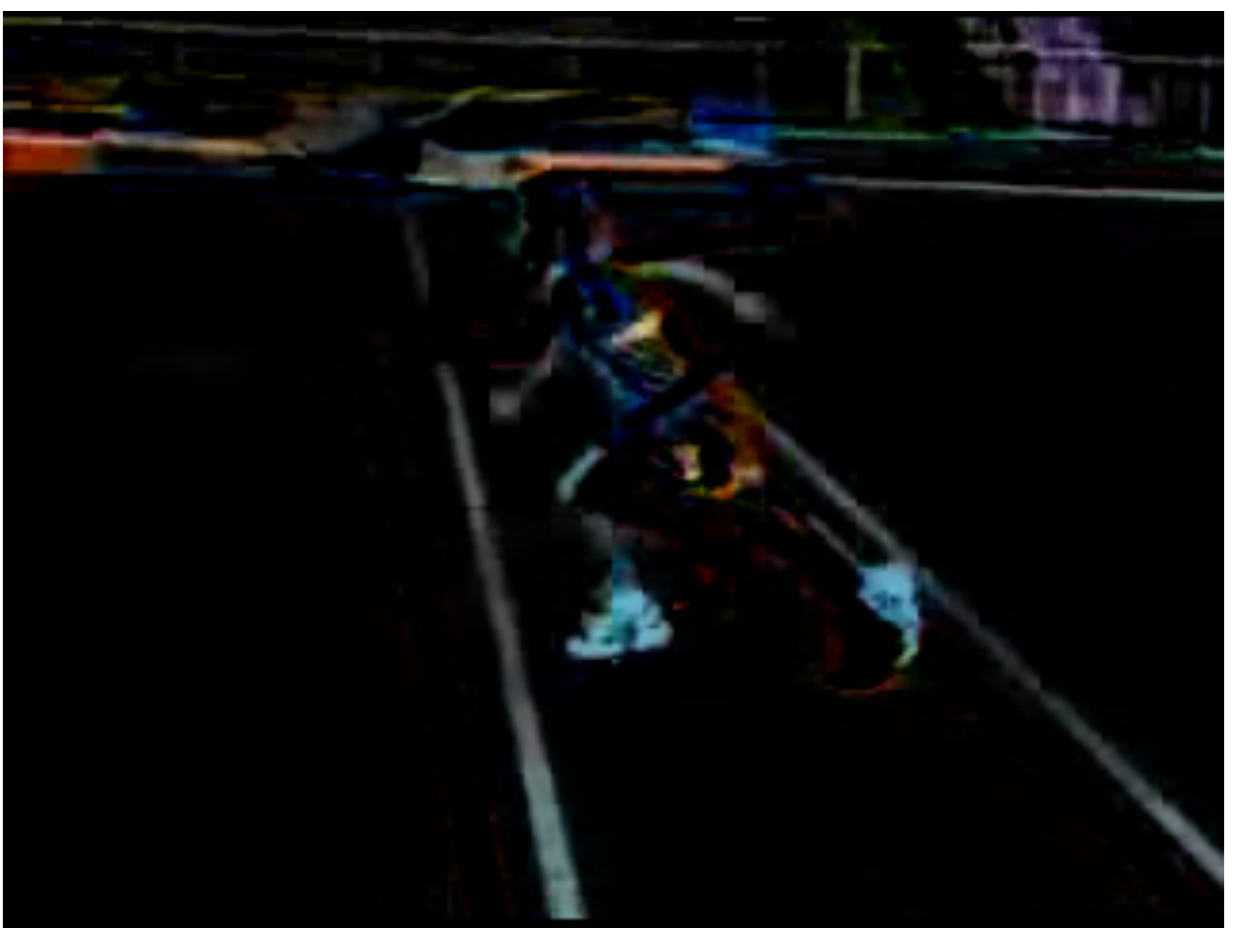}
		\includegraphics[width=0.2\linewidth]{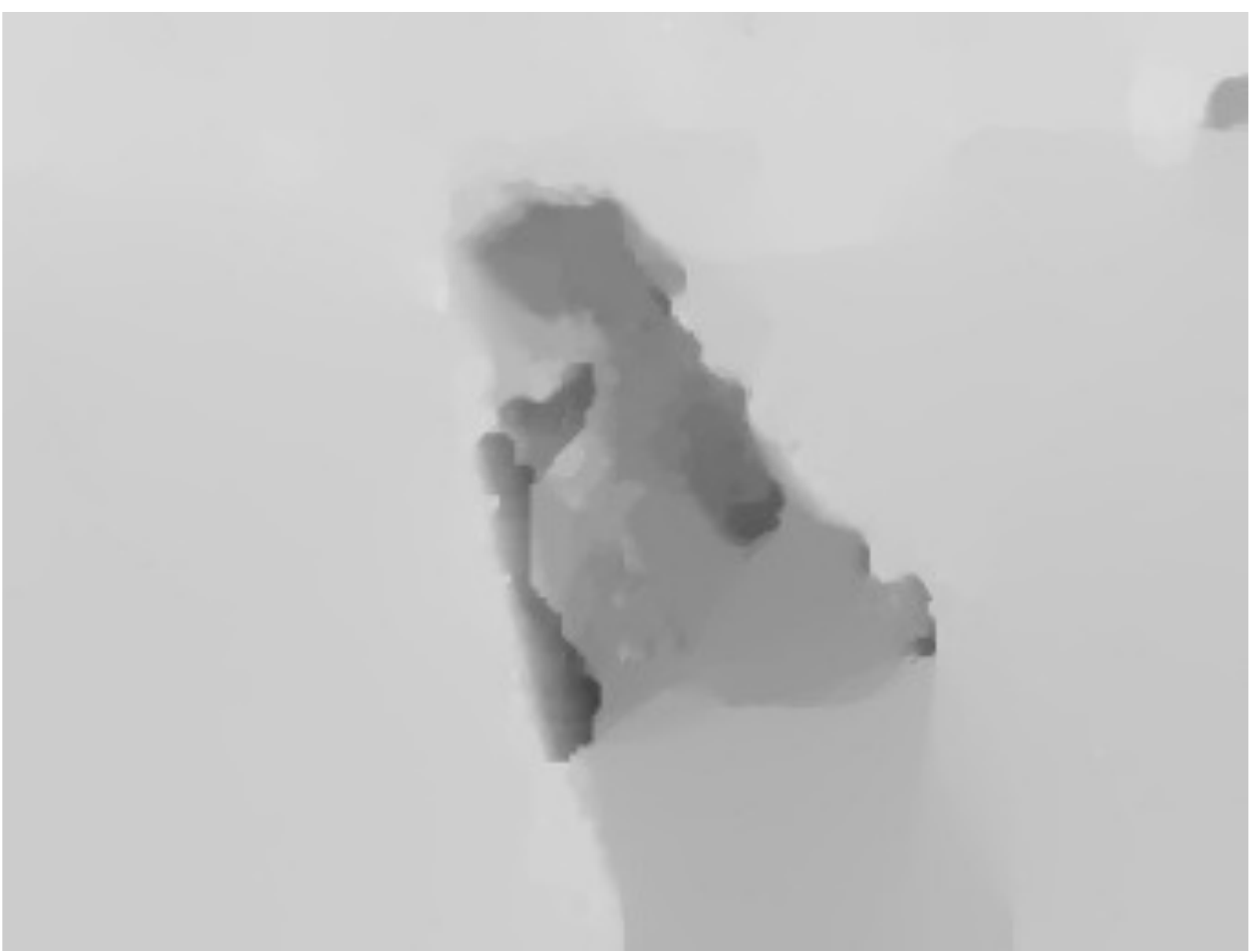}
		\includegraphics[width=0.2\linewidth]{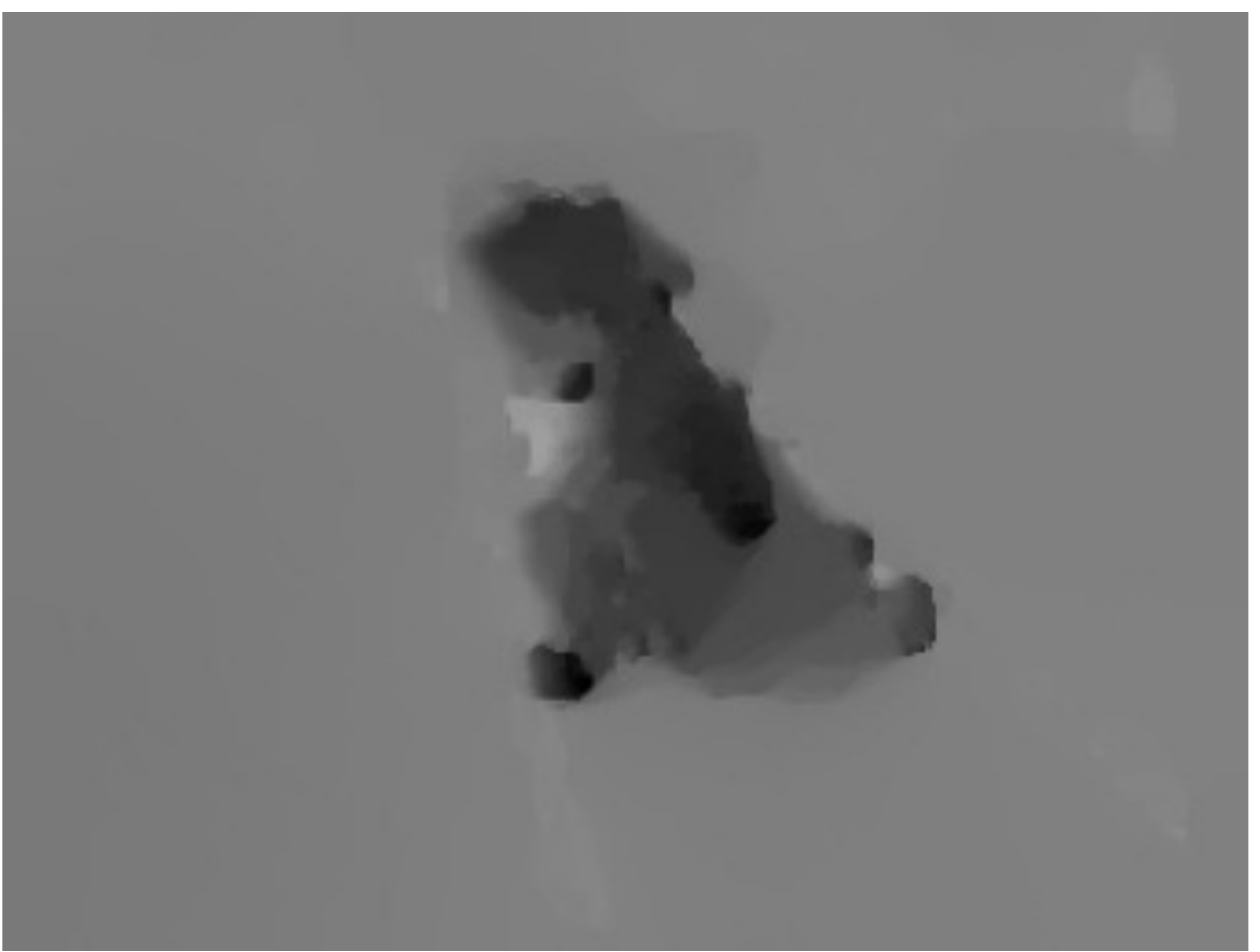} \\
		\includegraphics[width=0.2\linewidth]{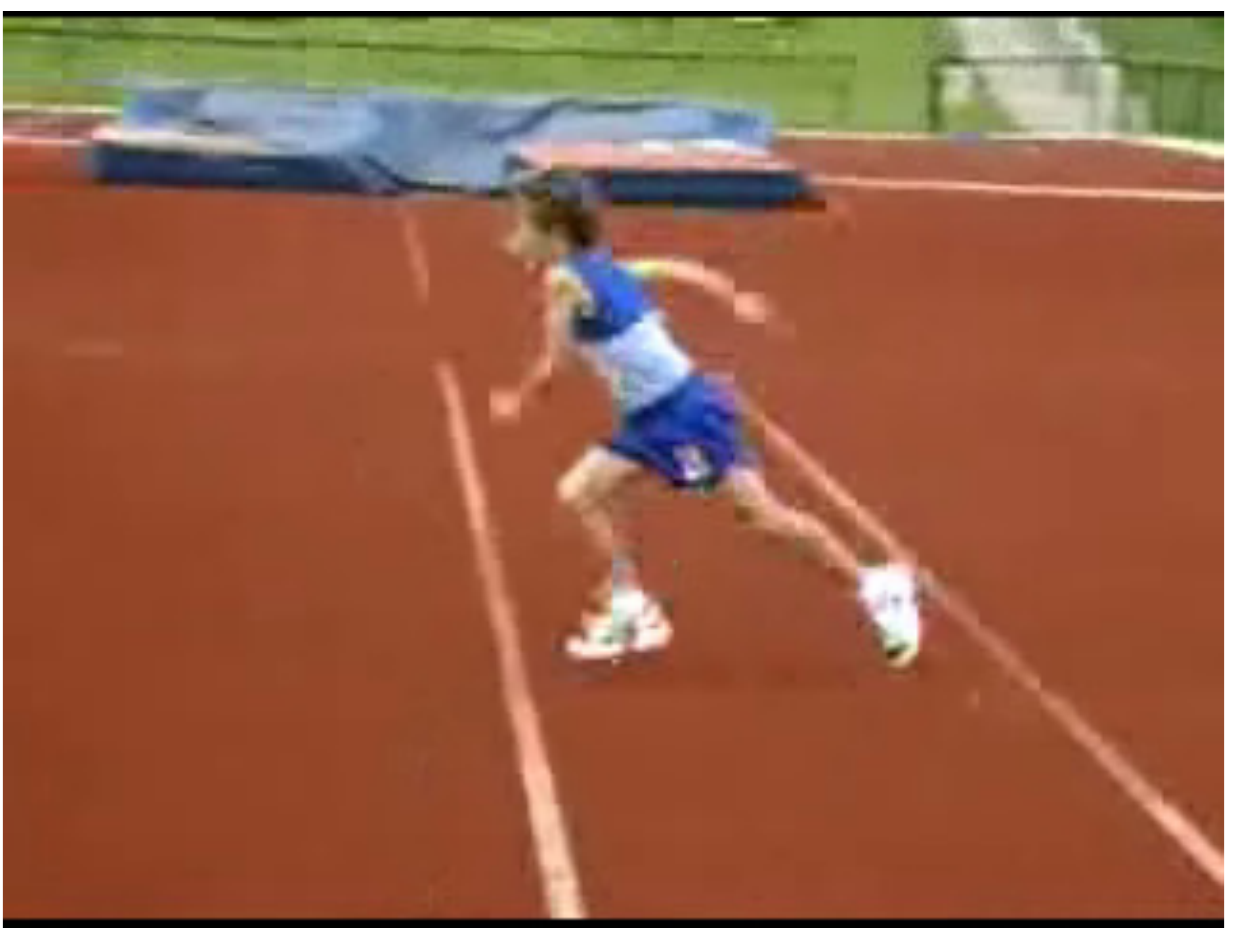}
		\includegraphics[width=0.2\linewidth]{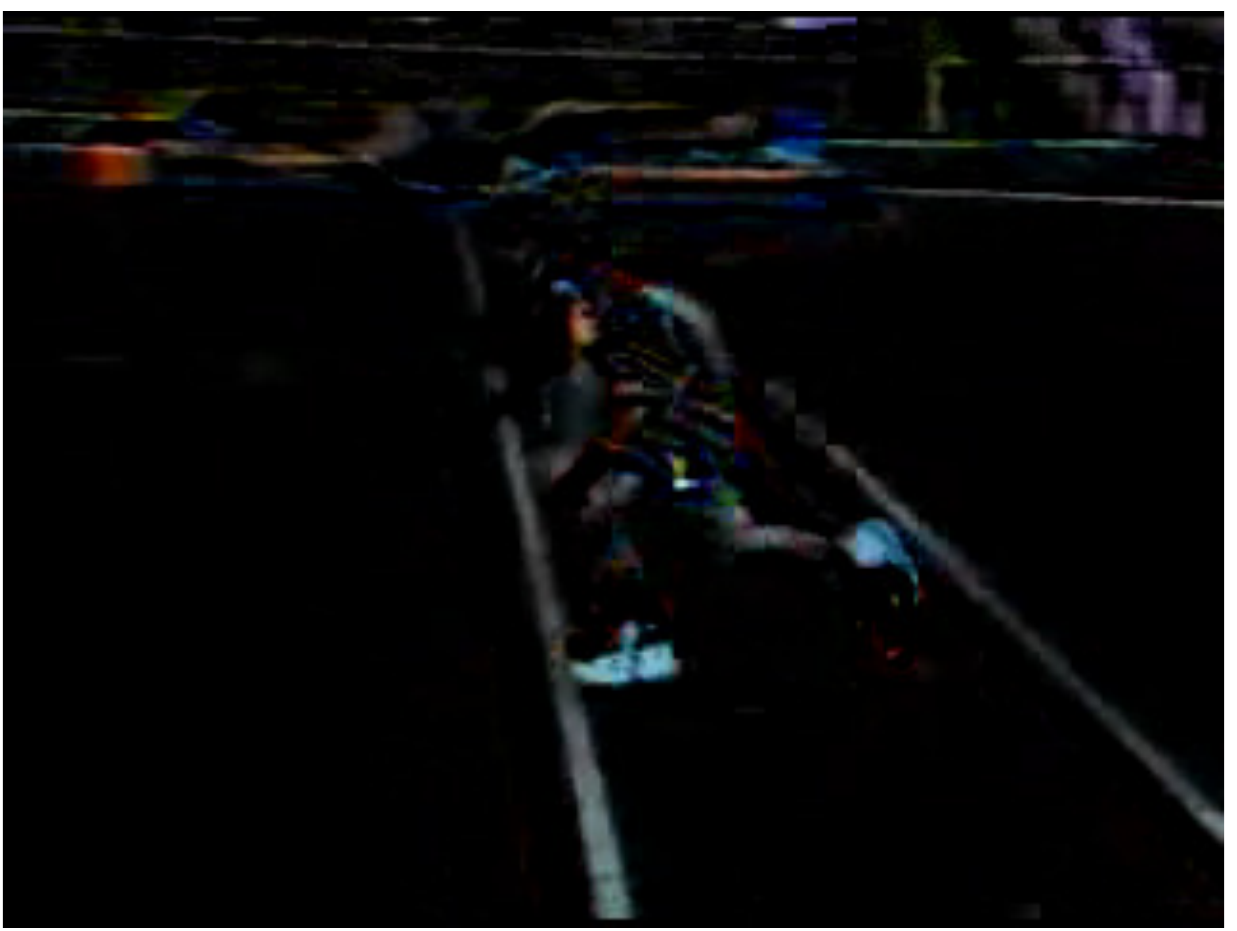}
		\includegraphics[width=0.2\linewidth]{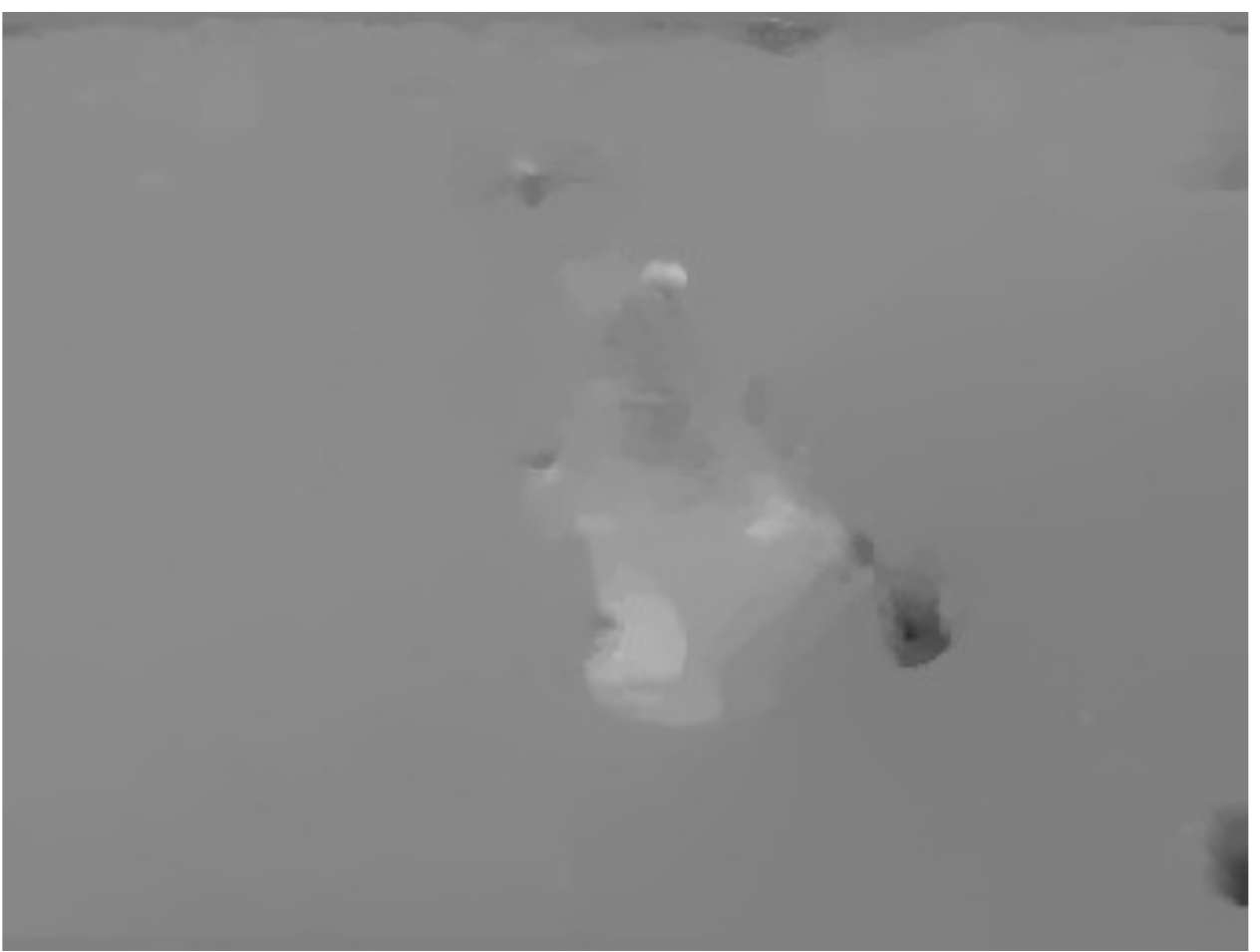}
		\includegraphics[width=0.2\linewidth]{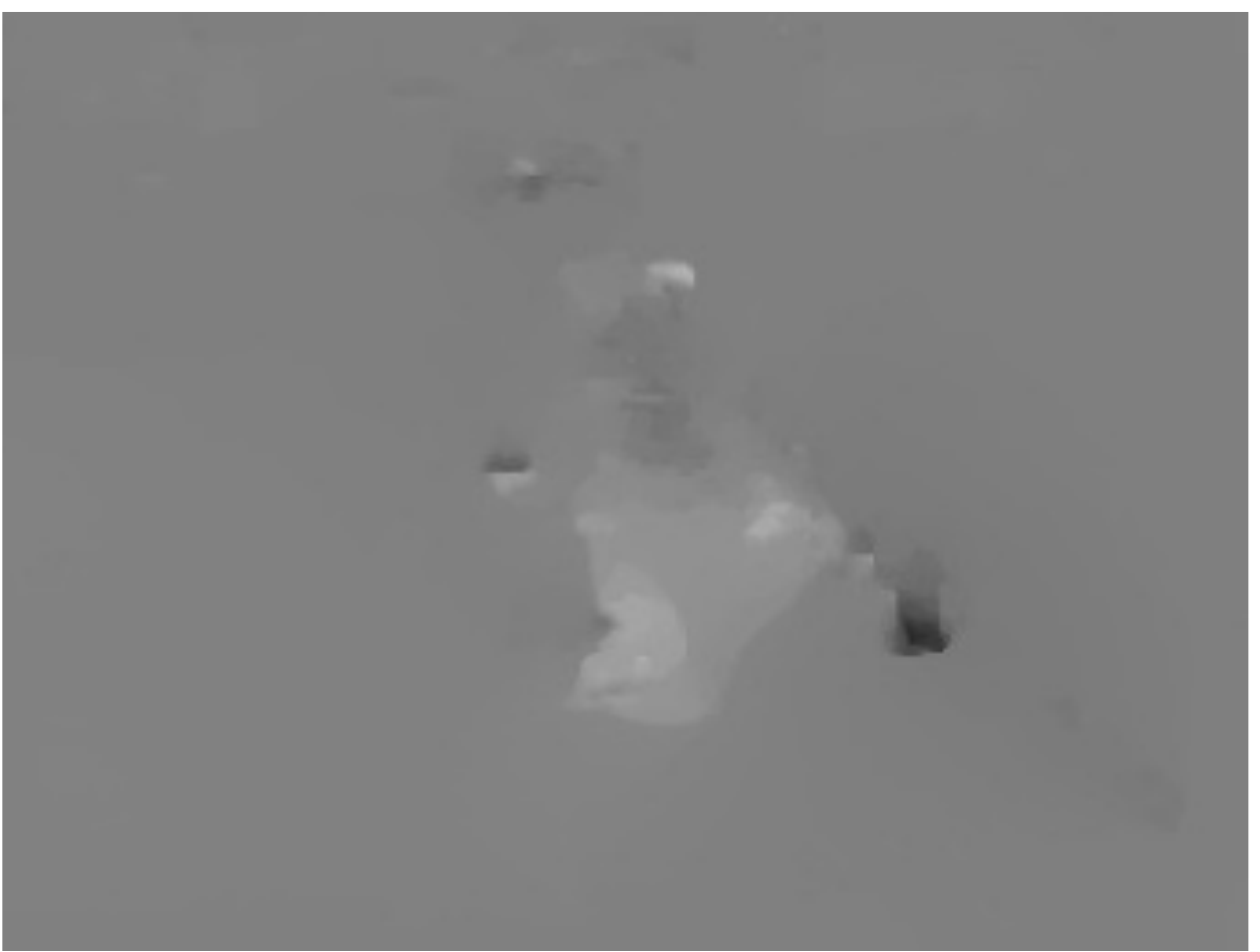} \\
	}
	\caption{Examples of four types of input modality: RGB images, RGB difference, optical flow fields (x,y directions), and warped optical flow fields (x,y directions)}
	\label{fig:ex}
\end{figure}

A single RGB image usually encodes static appearance at a specific time point and lacks the contextual information about previous and next frames. 
As shown in Figure \ref{fig:ex}, RGB difference between two consecutive frames describe the appearance change, which may correspond to the motion salient region. 
Inspired by~\cite{SunJYS15}, We experiment with adding stacked RGB difference as another input modality and investigate its performance in action recognition. 

The temporal stream ConvNets take optical flow field as input and aim to capture the motion information. In realistic videos, however, there usually exists camera motion, and optical flow fields may not concentrate on the human action. As shown in Figure \ref{fig:ex}, a remarkable amount of horizontal movement is highlighted in the background due to the camera motion. Inspired by the work of improved dense trajectories \cite{WangS13a}, we propose to take warped optical flow fields as additional input modality.
Following \cite{WangS13a}, we extract the warped optical flow by first estimating homography matrix and then compensating camera motion.
As shown in Figure \ref{fig:ex}, the warped optical flow suppresses the background motion and makes motion concentrate on the actor.

\textbf{Network Training.}
As the datasets for action recognition are relatively small, training deep ConvNets is challenged by the risk of over-fitting. 
To mitigate this problem, we design several strategies for training the ConvNets in \SEGNET s as follows.

\textit{Cross Modality Pre-training.} 
Pre-training has turned out to be an effective way to initialize deep ConvNets when the target dataset does not have enough training samples \cite{SimonyanZ14}. As spatial networks take RGB images as input, it is natural to exploit models trained on the ImageNet \cite{DengDSLL009} as initialization. For other modalities such as optical flow field and RGB difference, they essentially capture different visual aspects of video data and their distributions are different from that of RGB images. We come up with a cross modality pre-training technique in which we utilize RGB models to initialize the temporal networks. First, we discretize optical flow fields into the interval from 0 to 255 by a linear transformation. This step makes the range of optical flow fields to be the same with RGB images. Then, we modify the weights of first convolution layer of RGB models to handle the input of optical flow fields. Specifically, we average the weights across the RGB channels and replicate this average by the channel number of temporal network input. This initialization method works pretty well for temporal networks and reduce the effect of over-fitting in experiments.

\textit{Regularization Techniques.} 
Batch Normalization \cite{IoffeS15} is an important component to deal with the problem of covariate shift. In the learning process, batch normalization will estimate the activation mean and variance within each batch and use them to transform these activation values into a standard Gaussian distribution. This operation speeds up the convergence of training but also leads to over-fitting in the transferring process, due to the biased estimation of activation distributions from limited number of training samples. Therefore, after initialization with pre-trained models, we choose to freeze the mean and variance parameters of all Batch Normalization layers except the first one. As the distribution of optical flow is different from the RGB images, the activation value of first convolution layer will have a different distribution and we need to re-estimate the mean and variance accordingly. We call this strategy \textbf{partial BN}.
Meanwhile, we add a extra \textbf{dropout} layer after the global pooling layer in BN-Inception architecture to further reduce the effect of over-fitting. The dropout ratio is set as 0.8 for spatial stream ConvNets and $ 0.7 $ for temporal stream ConvNets.

\textit{Data Augmentation.} 
Data augmentation can generate diverse training samples and prevent severe over-fitting. In the original two-stream ConvNets, random cropping and horizontal flipping are employed to augment training samples. We exploit two new data augmentation techniques: corner cropping and scale-jittering. In corner cropping technique, the extracted regions are only selected from the corners or the center of the image to avoid implicitly focusing on the center area of a image. In multi-scale cropping technique, we adapt the scale jittering technique \cite{SimonyanZ14a} used in ImageNet classification to action recognition. We present an efficient implementation of scale jittering. We fix the size of input image or optical flow fields as $256 \times 340$, and the width and height of cropped region are randomly selected from $\{256, 224, 192, 168\}$. Finally, these cropped regions will be resized to $224 \times 224$ for network training. In fact, this implementation not only contains scale jittering, but also involves aspect ratio jittering.

\subsection{Testing Temporal Segment Networks}

Finally, we present our testing method for \SEGNET s. 
Due to the fact that all snippet-level ConvNets share the model parameters in \SEGNET s, the learned models can perform frame-wise evaluation as normal ConvNets.
This allows us to carry out fair comparison with models learned without the \SEGNET~framework. 
Specifically, we follow the testing scheme of the original two-stream ConvNets \cite{SimonyanZ14}, where we sample $ 25 $ RGB frames or optical flow stacks from the action videos. Meanwhile, we crop $ 4 $ corners and $ 1 $ center, and their horizontal flipping from the sampled frames to evaluate the ConvNets.  
For the fusion of spatial and temporal stream networks, we take a weighted average of them. 
When learned within the \SEGNET framework, the performance gap between spatial stream ConvNets and temporal stream ConvNets is much smaller than that in the original two-stream ConvNets.
Based on this fact, we give more credits to the spatial stream by setting its weight as $1$ and that of temporal stream as $1.5$.
When both normal and warped optical flow fields are used, the weight of temporal stream is divided to $ 1 $ for optical flow and $ 0.5 $ for warped optical flow.
It is described in Sec.~\ref{sec:seg} that the segmental consensus function is applied before the Softmax normalization.
To test the models in compliance with their training, we fuse the prediction scores of $ 25 $ frames and different streams before Softmax normalization.

\section{Experiments}
\label{sec:exp}

In this section, we first introduce the evaluation datasets and the implementation details of our approach. Then, we explore the proposed good practices for learning \SEGNET s. After this, we demonstrate the importance of modeling long-term temporal structures by applying the \SEGNET ~framework. We also compare the performance of our method with the state of the art. Finally, we visualize our learned ConvNet models.
\subsection{Datasets and Implementation Details}

We conduct experiments on two large action datasets, namely HMDB51 \cite{KuehneJGPS11} and UCF101 \cite{Soomro12}. The UCF101 dataset contains $ 101 $ action classes and $13,320$ video clips. We follow the evaluation scheme of the THUMOS13 challenge \cite{THUMOS13} and adopt the three training/testing splits for evaluation.  The HMDB51 dataset is a large collection of realistic videos from various sources, such as movies and web videos. The dataset is composed of $6,766$ video clips from $51$ action categories. Our experiments follow the original evaluation scheme using three training/testing splits and report average accuracy over these splits.

We use the mini-batch stochastic gradient descent algorithm to learn the network parameters, where the batch size is set to $ 256 $ and momentum set to $ 0.9 $. We initialize network weights with pre-trained models from ImageNet \cite{DengDSLL009}. We set a smaller learning rate in our experiments. For spatial networks, the learning rate is initialized as $0.001$ and decreases to its $\frac{1}{10}$ every $ 2,000 $ iterations. The whole training procedure stops at $ 4,500 $ iterations. For temporal networks, we initialize the learning rate as $0.005$, which reduces to its $\frac{1}{10}$ after $ 12,000 $ and $18,000$ iterations. The maximum iteration is set as $20,000$. Concerning data augmentation, we use the techniques of location jittering, horizontal flipping, corner cropping, and scale jittering, as specified in Section \ref{sec:training}. For the extraction of optical flow and warped optical flow, we choose the TVL1 optical flow algorithm \cite{ZachPB07} implemented in OpenCV with CUDA.
To speed up training, we employ a data-parallel strategy with multiple GPUs, implemented with our modified version of Caffe~\cite{JiaSDKLGGD14} and OpenMPI \footnote{\url{https://github.com/yjxiong/caffe}}. The whole training time on UCF101 is around 2 hours for spatial TSNs and 9 hours for temporal TSNs with 4 TITANX GPUs.

\subsection{Exploration Study}

In this section, we focus on the investigation the good practices described in Sec.~\ref{sec:training}, including the training strategies and the input modalities. In this exploration study, we use the two-stream ConvNets with very deep architecture adapted from~\cite{IoffeS15} and perform all experiments on the split 1 of UCF101 dataset.

We propose two training strategies in Section~\ref{sec:training}, namely cross modality pre-training and partial BN with dropout. Specifically, we compare four settings: (1) training from scratch, (2) only pre-train spatial stream as in~\cite{SimonyanZ14}, (3) with cross modality pre-training, (4) combination of cross modality pre-training and partial BN with dropout. The results are summarized in Table \ref{tbl:learning}. First, we see that the performance of training from scratch is much worse than that of the original two-stream ConvNets (baseline), which implies carefully designed learning strategy is necessary to reduce the risk of over-fitting, especially for spatial networks. Then, We resort to the pre-training of the spatial stream and cross modality pre-training of the temporal stream to help initialize two-stream ConvNets and it achieves better performance than the baseline. We further utilize the partial BN with dropout to regularize the training procedure, which boosts the recognition performance to $92.0\%$.

\begin{table}[t]
	\begin{center}
		\caption{Exploration of different training strategies for two-stream ConvNets on the UCF101 dataset (split 1).}
		\label{tbl:learning}
		\begin{tabular}{|l|c|c|c|}
			\hline
			Training setting & Spatial ConvNets  & Temporal ConvNets & Two-Stream \\
			\hline 
			Baseline \cite{SimonyanZ14} & $ 72.7\% $ & $ 81.0\% $ & $  87.0\%  $\\
			\hline
			From Scratch  &  $ 48.7\% $ &  $ 81.7\% $ & $ 82.9\% $ \\
			\hline
			Pre-train Spatial(same as~\cite{SimonyanZ14})  & $ 84.1\% $ &  $ 81.7\% $ & $ 90.0\% $ \\
			\hline
			+ Cross modality pre-training  & $ 84.1\% $ &  $ 86.6\% $ & $ 91.5\% $ \\
			\hline
			+ Partial BN with dropout & $ 84.5 \% $ & $ 87.2\% $ & $ 92.0\% $ \\
			\hline
		\end{tabular}
	\end{center}
\end{table}

We propose two new types of modalities in Section~\ref{sec:training}: RGB difference and warped optical flow fields. 
Results on comparing the performance of different modalities are reported in Table~\ref{tbl:modality}. 
These experiments are carried out with all the good practices verified in Table~\ref{tbl:learning}. We first observe that the combination of RGB images and RGB differences boosts the recognition performance to $ 87.3\% $ . This result indicates that RGB images and RGB difference may encode complementary information. Then it is shown that optical flow and warped optical flow yield quite similar performance ($ 87.2\% $ vs. $ 86.9\% $) and the fusion of them can improve the performance to $ 87.8\% $. Combining all of four modalities leads to an accuracy of $ 91.7\% $.  
As RGB difference may describe similar but unstable motion patterns, we also evaluate the performance of combining the other three modalities and this brings better recognition accuracy ($ 92.3\% $ vs $ 91.7\% $). 
We conjecture that the optical flow is better at capturing motion information and sometimes RGB difference may be unstable for describing motions. 
On the other hand, RGB difference may serve as a low-quality, high-speed alternative for motion representations.

\begin{table}[t]
	\begin{center}
		\caption{Exploration of different input modalities for two-stream ConvNets on the UCF101 dataset (split 1).}
		\label{tbl:modality}
		\begin{tabular}{|l|c|}
			\hline
			Modality & Performance \\
			\hline 
			RGB Image & $ 84.5\% $  \\
			\hline
			RGB Difference &  $ 83.8\% $ \\
			\hline
			RGB Image + RGB Difference & $ 87.3\% $ \\
			\hline \hline 
			Optical Flow  & $ 87.2\% $ \\
			\hline
			Warped Flow & $ 86.9\% $ \\
			\hline
			Optical Flow + Warped Flow & $ 87.8\% $ \\
			\hline \hline
			Optical Flow + Warped Flow + RGB & $\mathbf{92.3\%}$\\
			\hline
			All Modalities& $ 91.7\% $\\
			\hline
		\end{tabular}
	\end{center}
\end{table}

\begin{table}[t]
	\begin{center}
		\caption{Exploration of different segmental consensus functions for \SEGNET s on the UCF101 dataset (split 1).}
		\label{tbl:segmental}
		\begin{tabular}{|l|c|c|c|}
			\hline
			Consensus Function & Spatial ConvNets  & Temporal ConvNets & Two-Stream \\
			\hline 
			Max & $ 85.0\%  $& $ 86.0\% $ & $ 91.6\% $ \\
			\hline
			Average  &  $ 85.7\% $ &  $ 87.9\% $ & $ \mathbf{93.5\%} $ \\
			\hline
			Weighted Average & $ 86.2\% $ & $  87.7\% $ & $ 92.4\% $ \\
			\hline
		\end{tabular}
	\end{center}
\end{table}

\subsection{Evaluation of Temporal Segment Networks}

In this subsection, we focus on the study of the \SEGNET~framework. We first study the effect of segmental consensus function and then compare different ConvNet architectures on the split $ 1 $ of UCF101 dataset. For fair comparison, we only use RGB images and optical flow fields for input modalities in this exploration. As mentioned in Sec~\ref{sec:seg}, the number of segments $ K $ is set to $ 3 $.

In~Eq. (\ref{equ:model}), a segmental consensus function is defined by its aggregation function $ g $. Here we evaluate three candidates: (1) max pooling, (2) average pooling, (3) weighted average, for the form of $ g $. The experimental results are summarized in Table \ref{tbl:segmental}. We see that average pooling function achieves the best performance. So in the following experiments, we choose average pooling as the default aggregation function. 
Then we compare the performance of different network architectures and the results are summarized in Table \ref{tbl:structure}. Specifically, we compare three very deep architectures: BN-Inception~\cite{IoffeS15}, GoogLeNet \cite{SzegedyLJSRAEVR14}, and VGGNet-16 \cite{SimonyanZ14a}, all these architectures are trained with the good practices aforementioned. 
Among the compared architectures, the very deep two-stream ConvNets adapted from BN-Inception~\cite{IoffeS15} achieves the best accuracy of 92.0\%. 
This is in accordance with its better performance in the image classification task.
So we choose BN-Inception~\cite{IoffeS15} as the ConvNet architecture for \SEGNET s.

With all the design choices set, we now apply the \SEGNET~(TSN) to the action recognition. The result is illustrated in Table~\ref{tbl:structure}.
A component-wise analysis of the components in terms of the recognition accuracies is also presented in Table~\ref{tbl:component}.
We can see that \SEGNET is able to boost the performance of the model even when all the discussed good practices are applied.
This corroborates that modeling long-term temporal structures is crucial for better understanding of action in videos.
And it is achieved by \SEGNET s.

\begin{table}[t]
	\begin{center}
		\caption{Exploration of different very deep ConvNet architectures on the UCF101 dataset (split 1). 
			``BN-Inception+TSN'' refers to the setting where the \SEGNET framework is applied on top of the best performing BN-Inception~\cite{IoffeS15} architecture.}
		\label{tbl:structure}
		\begin{tabular}{|l|c|c|c|}
			\hline
			Training setting & Spatial ConvNets  & Temporal ConvNets & Two-Stream \\
			\hline 
			Clarifai~\cite{SimonyanZ14} & $ 72.7\% $ & $ 81.0\% $ & $ 87.0\% $ \\
			\hline
			GoogLeNet  & $ 77.1\% $ &  $ 83.9\% $ & $ 89.0\% $ \\
			\hline
			VGGNet-16 & $ 79.8\% $ &  $ 85.7\% $ & $ 90.9\% $ \\
			\hline
			BN-Inception & $ 84.5\% $  & $ 87.2\% $ & $ 92.0\% $ \\
			\hline
			\hline
			BN-Inception+TSN  & $ 85.7 \%$ & $ 87.9 \%$ & $ \mathbf{93.5\%} $ \\
			\hline
		\end{tabular}
	\end{center}
\end{table}

\begin{table}[t]
	\begin{center}
		\caption{Component analysis of the proposed method on the UCF101 dataset (split 1). 
			From left to right we add the components one by one.
			BN-Inception~\cite{IoffeS15} is used as the ConvNet architecture.}
		\label{tbl:component}
		\begin{tabular}{|l|c|c|c|c|}
			\hline
			Component &\vtop{\hbox{\strut Basic}\hbox{\strut Two-Stream~\cite{SimonyanZ14}}}
					&\vtop{\hbox{\strut Cross-Modality}\hbox{\strut Pre-training}}
					&\vtop{\hbox{\strut Partial BN}\hbox{\strut with dropout}}
					&\vtop{\hbox{\strut Temporal}\hbox{\strut Segment Networks}}\\
			\hline
			Accuracy  & $ 90.0\% $ & $ 91.5 $ & $ 92.0\% $ & $ 93.5\% $
			\\\hline
		\end{tabular}
	\end{center}
\end{table}

\subsection{Comparison with the State of the Art}
\begin{table}[h]
	\small
	\centering
	\caption{Comparison of our method based on \SEGNET (TSN) with other state-of-the-art methods. We separately present the results of using two input modalities (RGB+Flow) and three input modalities (RGB+Flow+Warped Flow).}
	\begin{tabular}{|lr|lr|}
		\hline
		\multicolumn{2}{|c|}{HMDB51} & \multicolumn{2}{|c|}{UCF101} \\
		\hline
		\hline
		DT+MVSV \cite{CaiWPQ14} & $ 55.9\% $ & DT+MVSV \cite{CaiWPQ14} & $ 83.5\% $ \\
		iDT+FV \cite{WangS13a} & $ 57.2\% $ & iDT+FV \cite{WangS13b} & $ 85.9\% $ \\
		iDT+HSV \cite{PengWWQ14} & $ 61.1\% $ & iDT+HSV \cite{PengWWQ14} & $ 87.9\% $ \\
		MoFAP \cite{WangQT15b} & $ 61.7\% $ & MoFAP \cite{WangQT15b} & $ 88.3\% $ \\ 
		\hline
		\hline
		Two Stream \cite{SimonyanZ14} & $ 59.4\% $ & Two Stream \cite{SimonyanZ14} & $ 88.0\% $ \\
		VideoDarwin \cite{FernandoGMGT15} & $ 63.7\% $ & C3D (3 nets) \cite{TranBFTP15} & $ 85.2\% $ \\
		MPR \cite{NiMYY15} & $ 65.5\% $ & Two stream +LSTM \cite{Ng15} & $ 88.6\% $ \\
		$\mathrm{F_{ST}CN}$ (SCI fusion) \cite{SunJYS15} & $ 59.1\% $ & $\mathrm{F_{ST}CN}$ (SCI fusion) \cite{SunJYS15} & $ 88.1\% $ \\
		
		TDD+FV \cite{WangQT15a} & $ 63.2\% $ & TDD+FV \cite{WangQT15a} & $ 90.3\% $ \\
		LTC~\cite{varol} & $ 64.8\% $ & LTC~\cite{varol} & $ 91.7\% $ \\
		KVMF~\cite{ZhuW2016} & $ 63.3\% $ & KVMF~\cite{ZhuW2016} & $ 93.1\% $ \\
		\hline
		\hline
		TSN (2 modalities)~~~~~~~~~~~~~~~~ & $ 68.5\% $ & TSN (2 modalities)~~~~~~~~~~~~~~~~ & $ 94.0\% $  \\
		TSN (3 modalities) & $ \mathbf{69.4\%} $ & TSN (3 modalities) & $ \mathbf{94.2\%} $\\
		\hline
	\end{tabular}
	\label{tbl:stoa}
\end{table}

After exploring of the good practices and understanding the effect of \SEGNET, we are ready to build up our final action recognition method. Specifically, we assemble three input modalities and all the techniques described as our final recognition approach, and test it on two challenging datasets: HMDB51 and UCF101. The results are summarized in Table \ref{tbl:stoa}, where we compare our method with both traditional approaches such as improved trajectories (iDTs)~\cite{WangS13a}, MoFAP representations \cite{WangQT15b}, and deep learning representations, such as 3D convolutional networks (C3D) \cite{TranBFTP15}, trajectory-pooled deep-convolutional descriptors (TDD) \cite{WangQT15a}, factorized spatio-temporal convolutional networks ($\mathrm{F_{ST}CN}$) \cite{SunJYS15}, long term convolution networks (LTC)~\cite{varol}, and key volume mining framework (KVMF)~\cite{ZhuW2016}. Our best result outperforms other methods by $  3.9\% $ on the HMDB51 dataset, and $ 1.1\% $ on the UCF101 dataset. The superior performance of our methods demonstrates the effectiveness of \SEGNET and justifies the importance of long-term temporal modeling.

\subsection{Model Visualization}
\begin{figure}[t]
	\centering
	\includegraphics[width=\linewidth]{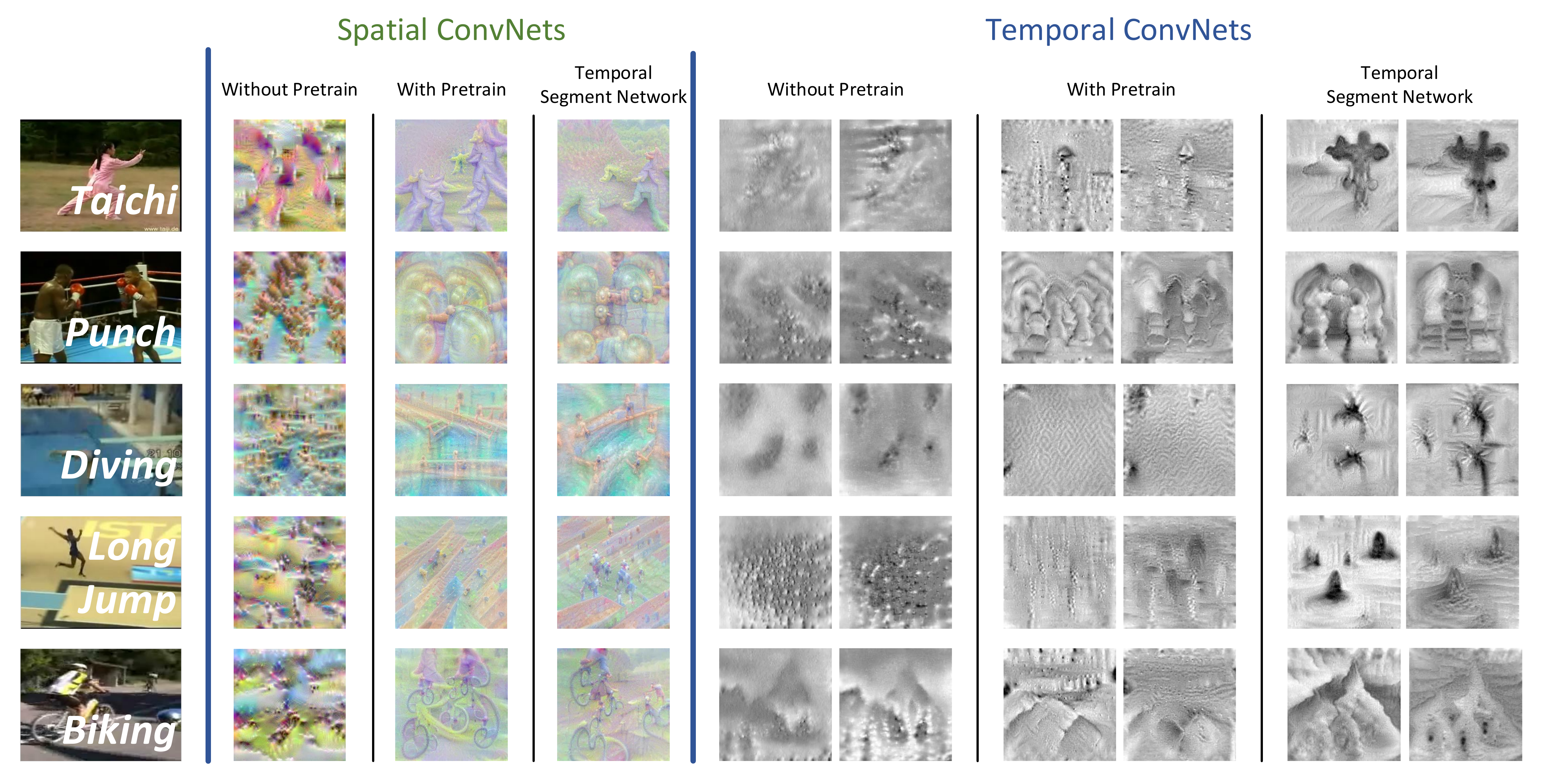}
	\caption{Visualization of ConvNet models for action recognition using DeepDraw~\cite{DeepDraw}. 
		We compare three settings: (1) without pre-train; (2) with pre-train; (3) with \SEGNET .
		For spatial ConvNets, we plot three generated visualization as color images.
		For temporal ConvNets, we plot the flow maps of $ x $ (left) and $ y $ (right) directions in gray-scales.
		Note all these images are generated from purely random pixels.}
	\label{fig:visualization}
\end{figure}

Besides recognition accuracies, we would like to attain further insight into the learned ConvNet models. In this sense, we adopt the DeepDraw~\cite{DeepDraw} toolbox. This tool conducts iterative gradient ascent on input images with only white noises. Thus the output after a number of iterations can be considered as class visualization based solely on class knowledge inside the ConvNet model. The original version of the tool only deals with RGB data. To conduct visualization on optical flow based models, we adapt the tool to work with our temporal ConvNets. As a result, we for the first time visualize interesting class information in action recognition ConvNet models. We randomly pick five classes from the UCF101 dataset,~\emph{Taichi},~\emph{Punch},~\emph{Diving},~\emph{Long Jump}, and~\emph{Biking} for visualization. The results are shown in Fig.~\ref{fig:visualization}. For both RGB and optical flow, we visualize the ConvNet models learned with following three settings: (1) without pre-training; (2) only with pre-training; (3) with \SEGNET.

Generally speaking, models with pre-training are more capable of representing visual concepts than those without pre-training. One can see that both spatial and temporal models without pre-training can barely generate any meaningful visual structure. With the knowledge transferred from the pre-training process, the spatial and temporal models are able to capture structured visual patterns.

It is also easy to notice that the models, trained with only short-term information such as single frames, tend to mistake the scenery patterns and objects in the videos as significant evidences for action recognition. 
For example, in the class ``Diving'', the single-frame spatial stream ConvNet mainly looks for water and diving platforms, other than the person performing diving. Its temporal stream counterpart, working on optical flow, tends to focus on the motion caused by waves of surface water. 
With long-term temporal modeling introduced by \SEGNET , it becomes obvious that learned models focus more on humans in the videos, and seem to be modeling the long-range structure of the action class. Still consider ``Diving'' as the example, the spatial ConvNet with \SEGNET now generate a image that human is the major visual information. And different poses can be identified in the image, depicting various stages of one diving action. 
This suggests that models learned with the proposed method may perform better, 
which is well reflected in our quantitative experiments. 
We refer the reader to supplementary materials for visualization of more action classes and more details on the visualization process.

\section{Conclusions}
\label{sec:con}

In this paper, we presented the Temporal Segment Network (TSN), a video-level framework that aims to model long-term temporal structure. As demonstrated on two challenging datasets, this work has brought the state of the art to a new level, while maintaining a reasonable computational cost. This is largely ascribed to the segmental architecture with sparse sampling, as well as a series of good practices that we explored in this work. The former provides an effective and efficient way to capture long-term temporal structure, while the latter makes it possible to train very deep networks on a limited training set without severe overfitting.\\

\noindent \textbf{Acknowledgment.}
This work was supported by the \emph{Big Data Collaboration Research} grant from SenseTime Group (CUHK Agreement No. TS1610626), Early Career Scheme (ECS) grant (No. 24204215), ERC Advanced Grant {\em VarCity} (No. 273940), Guangdong Innovative Research Program (2015B010129013, 2014B050 505017), and Shenzhen Research Program (KQCX2015033117354153, JSGG2015 0925164740726, CXZZ20150930104115529), and External Cooperation Program of BIC, Chinese Academy of Sciences (172644KYSB20150019).

\clearpage

\bibliographystyle{splncs}
\bibliography{deep}

\end{document}